# The 'Letter' Distribution in the Chinese Language

Qinghua Chen[abc*], Yan Wang[d], Mengmeng Wang[e] and Xiaomeng Li[a*]

[a]School of Systems Science, Beijing Normal University, Beijing 100875, China;

[b]New England Complex Systems Institute, Cambridge, MA 02139, USA;

[c]Department of Chemistry, Brandeis University, Waltham, MA 02453, USA;

[d]Department of Mathematics, University of California, Los Angeles 90095, USA;

[e]Business College, Shanghai Normal University, Shanghai 200234, China

*Corresponding authors: qinghuachen@bnu.edu.cn, lixiaomeng@bnu.edu.cn

**ABSTRACT**
Corpus-based statistical analysis plays a significant role in linguistic research, and ample evidence has shown that different languages exhibit some common laws. Studies have found that letters in some alphabetic writing languages have strikingly similar statistical usage frequency distributions. Does this hold for Chinese, which employs ideogram writing? We obtained letter frequency data of some alphabetic writing languages and found the common law of the letter distributions. In addition, we collected Chinese literature corpora for different historical periods from the Tang Dynasty to the present, and we dismantled the Chinese written language into three kinds of basic particles: characters, strokes and constructive parts. The results of the statistical analysis showed that, in different historical periods, the intensity of the use of basic particles in Chinese writing varied, but the form of the distribution was consistent. In particular, the distributions of the Chinese constructive parts are certainly consistent with those alphabetic writing languages. This study provides new evidence of the consistency of human languages.

**KEYWORDS**
letter distribution of language, Chinese constructive part, Zipf's plot, KS statistics, common law

## 1. Introduction

Language has a fundamentally social function. As a human-driven complex adaptive system (Liu, 2018), language has attracted much interest from researchers. In the study of language, corpus-based statistical analysis has played a significant role (Manning, Manning, & Schütze, 1999). Some experimental discoveries inspire scholars proposed that there might be some "common laws" for different languages. A representative pioneer was George Kingsley Zipf, who found that the word frequency distributions of some human languages follow Zipf's law (Zipf, 1949), which is also verified by other scholars later, as Korean (Choi, 2000), Greek (Hatzigeorgiu, Mikros, & Carayannis, 2001), Turkish (Dalkılıç & Çebi, 2004), French and Spanish (Ha, Stewart, Hanna, & Smith, 2006), some Indian languages (Jayaram & Vidya, 2008), Arabic (Masrai & Milton, 2016) and German, Latin, Afrikaans, Indonesian, Somali (Wiegand, Nadarajah, & Si, 2018). Besides these, Chinese, as a representative of ideograms, the word using always shows Zipf's law in different periods including Tang Dynasty, Song Dynasty, Yuan Dynasty, Ming Dynasty, Qing Dynasty and present based on statistics Chinese (Q. Chen, Guo, & Liu, 2012). Esperanto, as a constructed language, also follow this universal law (Wiegand et al., 2018). Because of this amazing consistency, the empirical studies on word usage frequencies have been a focus point of studies in statistical linguistics for the past 70 years (Piantadosi, 2014).

Besides word usage frequencies, some other statistical work has been carried out which also focuses on searching the "common law" for various languages. Heaps found that the number of different words (i.e., word types) scales with database size measured in the total number of words in various languages (Heaps, 1978). Based on the word ngram model, Brown et al. have attributed words to different classes through a statistical algorithm and have found different languages to be consistent in their semantics hierarchy (Brown, Desouza, Mercer, Pietra, & Lai, 1992). Using complex network theory, the scientists have found different syntactic dependency networks share many nontrivial statistical patterns such as the small-world phenomenon, scaling in the distribution of degrees (i Cancho, Solé, & Köhler, 2004; Liu, 2008). Zhang et al. used word2vec, which can convert words to vectors, to handle the conversion of text into vector space operations, he has used the similarity of the vector space to represent semantic similarities and has found different language structures to be consistent (Zhang, Xu, Su, & Xu, 2015). Dodds et al. have analysed the most commonly used words of 24 corpora across 10 diverse human languages and have found clear positive bias for all corpora (Dodds et al., 2015). Youn collected materials in different languages to explore the frequencies of polysemy, which represents the same concept, to measure their semantic closeness, and found that structural features are the same in different languages (Youn et al., 2016). Furthermore, some common motifs have been concentrated and found in different types of languages. (Beliankou, Köhler, & Naumann, 2012; Jing & Liu, 2017)

These "common laws" for different languages prompt the following scientific question: is the nature of language the same? At present, it is difficult to provide a convincing answer, and related research is far from adequate. Scholars need to continue working on the following two aspects: whether there is more evidence that might indicate that language is consistent and whether any theories might demonstrate the rationality of this consistency.

In this paper, we attempted to provide more evidence of the consistency of language. We focused on letters, which are the basic orthographic units of alphabetic writing languages. Scholars have found some common basic units of alphabetic writing language. In particular, letters have strikingly similar statistical distributions, for example in Spanish and English, as described in references (Jernigan, 2008; Li & Miramontes, 2011). The letter frequency distribution in the Voynich manuscript was analysed and found to be very similar to Moldavian, Karakalpak, Kabardian Circassian, Kannada, and Thai (Jaskiewicz, 2011). To what extent does this rule apply to other languages? In particular, does the same law exist in Chinese? Since there are no explicit letters in Chinese, thinking about this problem is challenging.

By using the frequency data of ten alphabetic languages from Wikipedia, we compared their distributions and found that the frequency distribution was consistent. Furthermore, we constructed a Chinese corpus from the literature of different historical periods and attempted to identify Chinese "letters". The article structure arranged as follows. In Section 2, we introduce the data sources used in this paper. In Section 3, we compare the letter distributions of the 10 alphabetic writing languages and find the best fitting curve for the letter frequencies. In Section 4, we discuss 3 possible candidates for Chinese "letters", including Chinese characters, basic strokes, and constructive parts. Their distributions are calculated and compared with the discussion on alphabetic languages. We conclude the paper in the last section.

## 2. Data sources

### 2.1. *Data for 10 alphabetic writing languages*

We obtained the frequency of letters of several alphabetic writing languages via the Wikipedia website https://en.wikipedia.org/wiki/Letter_frequency. The data encompass ten languages, including English, French, German, Spanish, Portuguese, Italian, Turkish, Swedish, Polish, and Esperanto, which is an artificially constructed language. However, They use rather similar letters.

**Table 1.** Letters and their relative usage frequencies (%).

|   | English | French | German | Spanish | Portuguese | Italian | Turkish** | Swedish | Polish | Esperanto |
|---|---------|--------|--------|---------|------------|---------|-----------|---------|--------|-----------|
| a | 8.167 | 7.636 | 6.516 | 11.525 | 14.634* | 11.745 | 12.92* | 9.383 | 10.503* | 12.117* |
| b | 1.492 | 0.901 | 1.886 | 2.215 | 1.043 | 0.927 | 2.844 | 1.535 | 1.74 | 0.98 |
| c | 2.782 | 3.26 | 2.732 | 4.019 | 3.882 | 4.501 | 1.463 | 1.486 | 3.895 | 0.776 |
| d | 4.253 | 3.669 | 5.076 | 5.01 | 4.992 | 3.736 | 5.206 | 4.702 | 3.725 | 3.044 |
| e | 12.702* | 14.715* | 16.396* | 12.181* | 12.57 | 11.792* | 9.912 | 10.149* | 7.352 | 8.995 |
| f | 2.228 | 1.066 | 1.656 | 0.692 | 1.023 | 1.153 | 0.461 | 2.027 | 0.143 | 1.037 |
| g | 2.015 | 0.866 | 3.009 | 1.768 | 1.303 | 1.644 | 1.253 | 2.862 | 1.731 | 1.171 |
| h | 6.094 | 0.737 | 4.577 | 0.703 | 0.781 | 0.636 | 1.212 | 2.09 | 1.015 | 0.384 |
| i | 6.966 | 7.529 | 6.55 | 6.247 | 6.186 | 10.143 | 9.6 | 5.817 | 8.328 | 10.012 |
| j | 0.153 | 0.613 | 0.268 | 0.493 | 0.397 | 0.011 | 0.034 | 0.614 | 1.836 | 3.501 |
| k | 0.772 | 0.074 | 1.417 | 0.011 | 0.015 | 0.009 | 5.683 | 3.14 | 2.753 | 4.163 |
| l | 4.025 | 5.456 | 3.437 | 4.967 | 2.779 | 6.51 | 5.922 | 5.275 | 2.564 | 6.104 |
| m | 2.406 | 2.968 | 2.534 | 3.157 | 4.738 | 2.512 | 3.752 | 3.471 | 2.515 | 2.994 |
| n | 6.749 | 7.095 | 9.776 | 6.712 | 4.446 | 6.883 | 7.987 | 8.542 | 6.237 | 7.955 |
| o | 7.507 | 5.796 | 2.594 | 8.683 | 9.735 | 9.832 | 2.976 | 4.482 | 6.667 | 8.779 |
| p | 1.929 | 2.521 | 0.67 | 2.51 | 2.523 | 3.056 | 0.886 | 1.839 | 2.445 | 2.755 |
| q | 0.095 | 1.362 | 0.018 | 0.877 | 1.204 | 0.505 | 0 | 0.02 | 0 | 0 |
| r | 5.987 | 6.693 | 7.003 | 6.871 | 6.53 | 6.367 | 7.722 | 8.431 | 5.243 | 5.914 |
| s | 6.327 | 7.948 | 7.27 | 7.977 | 6.805 | 4.981 | 3.014 | 6.59 | 5.224 | 6.092 |
| t | 9.056 | 7.244 | 6.154 | 4.632 | 4.336 | 5.623 | 3.314 | 7.691 | 2.475 | 5.276 |
| u | 2.758 | 6.311 | 4.166 | 2.927 | 3.639 | 3.011 | 3.235 | 1.919 | 2.062 | 3.183 |
| v | 0.978 | 1.838 | 0.846 | 1.138 | 1.575 | 2.097 | 0.959 | 2.415 | 0.012 | 1.904 |
| w | 2.36 | 0.049 | 1.921 | 0.017 | 0.037 | 0.033 | 0 | 0.142 | 5.813 | 0 |
| x | 0.15 | 0.427 | 0.034 | 0.215 | 0.253 | 0.003 | 0 | 0.159 | 0.004 | 0 |
| y | 1.974 | 0.128 | 0.039 | 1.008 | 0.006 | 0.02 | 3.336 | 0.708 | 3.206 | 0 |
| z | 0.074 | 0.326 | 1.134 | 0.467 | 0.47 | 1.181 | 1.5 | 0.07 | 4.852 | 0.494 |
| others | 0 | 2.832 | 2.323 | 2.978 | 4.138 | 1.292 | 6.692 | 4.44 | 7.687 | 2.33 |

[1]This table is incomplete with the exception of English and Dutch because other languages use other letters. For more specific information, refer to `https://en.wikipedia.org/wiki/Letter_frequency`.
[2]0 indicates that this letter does not exist in the language.
[3]Others denote frequency sum for other letters except for English letters. For example, there are è, é, ë and ê in French.
[*]This denotes highest frequency of letter in the language.
[**]There are inconsistencies in the Wiki data for Turkish. We have found the corresponding reference (Serengil & Akin, 2011) and made some corrections.

Here, we list the 26 most commonly used Latin letters across the languages in Table 1. This table clearly shows that the usage frequency has high variability among letters.
For example, in English, "e" is the most frequently used letter, and it is used almost 172 times more often than "z", which is the most rarely used letter in English. However, in the case of Esperanto, the proportion of use intensity of "e" and "z" is around
18. Among these languages, either "e" or "a" is the most frequently used; their usage frequency ranges from 10.149% to 16.396%.

The detailed frequency analysis is described in Section 3.

## 2.2. *Chinese data sources*

According to the reference (Q. Chen et al., 2012), the frequency of Chinese characters has changed significantly with the evolution of history. But the distribution for most of the words' using-frequency is much stable. However, the top list are different in different periods. The most popular words are "不", "春", "云", "了" and "的" respectively from Tang dynasty to present. To discuss the "letters" in Chinese comprehensively, we used 5 corpora in this paper, as in reference (Q. Chen et al., 2012). The data cover a wide range of Chinese literature. For convenience, we considered corpora from the Tang Dynasty to the 21st century, during the time that characters remained nearly the same. These materials were obtained from Internet sources, including http://www.tianyabook.com/. These corpora are described below. All of the materials are presented in simplified Chinese.

- Corpus 1 (618–907 A.D.):

    The Complete Tang poems.

- Corpus 2 (960–1279 A.D.):

    The Complete Song Ci-Poetry.

- Corpus 3 (1271–1368 A.D.):

    The Complete Yuan verse.

- Corpus 4 (1368–1911 A.D.):

    Four classical novels from the Ming and Qing Dynasties, viz. Story of a Journey to the West, All Men Are Brothers, Romance of the Three Kingdoms and Dream of the Red Chamber.

- Corpus 5 (after 2000 A.D.):

    Novels collected from the Internet, viz. eight stories from the most popular network story list (http://www.google.cn/rebang/) on April 20, 2009.

We only focused on Chinese characters and words, and we deleted all non-Chinese symbols, including punctuation marks, etc. The character counts are listed in Table 2. Further processing and analysis of the data are described in Section 4.

**Table 2.** The counts of characters per corpora.

|  | corpus 1 | corpus 2 | corpus 3 | corpus 4 | corpus 5 |
| --- | --- | --- | --- | --- | --- |
| count of characters | 2,602,310 | 1,417,778 | 2,172,631 | 2,506,684 | 12,379,116 |
| count of character type | 7,444 | 5,794 | 6,119 | 5,458 | 5,671 |

## 3. Letter distributions of alphabetic writing languages

### 3.1. *Letter distributions*

The letters are the natural units of the phonetic alphabet. Words are composed of one or more letters in a particular order. In English, the letter "e" is the most frequently used letter, at a rate of 12.702%. The letter with the lowest frequency of usage is "z", with a frequency of only 0.074%. In Portuguese, the most frequently used letter is "a", accounting for 14.634%, and the letter with the lowest frequency is "y", accounting for only 0.006%. Although Esperanto is a constructed language, it still have similar characters as natural languages in many fields (Ausloos, 2010; Wiegand et al., 2018). Here "a" is the most popular letter and it takes beyond 12% of the whole probability. These results reflect the different intensities of the usage of letters among different languages, but the letter frequencies share some certain similarities. In Figure 1, a Zipf's plot shows the correlation between the ranking and frequency of different letters for ten different languages in rank-size scale.

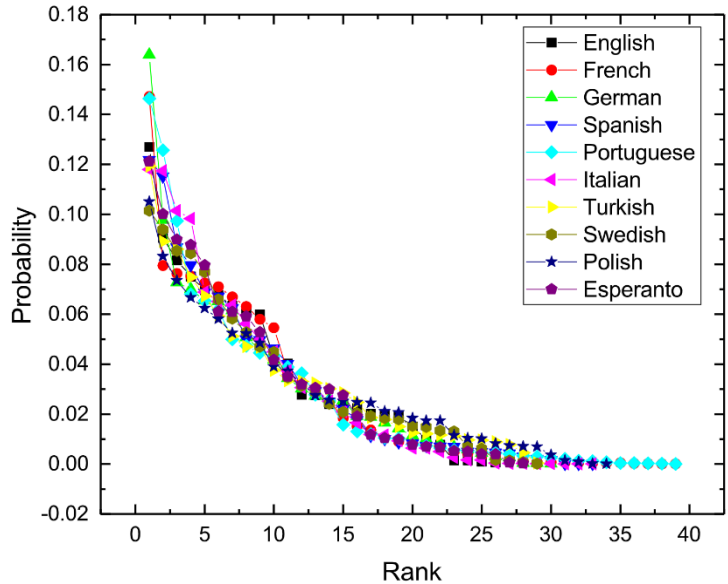

**Figure 1.** Letter frequency distributions for 10 alphabetical languages.

As shown in Figure 1, the distribution curves are monotonically decreasing. Even the curve for German and Portuguese are the steepest, the curve for Polish and Swedish are the least steep. In general, the decreasing curves are similar for all 10 languages.

Direct rough comparison is not easy to persuade. It should be noted that the number of letter types used in the 10 languages is different. To better compare usage, the letter ranking measurement should be re-scaled to be the proportion of the original rank to the total number of letters used, which always should be constrained between 0 and 1. The re-scaled rank-size plot is shown in Figure 2. These probability curves show a similar downward trend of fold change, which is the same result as that shown in Figure 1.

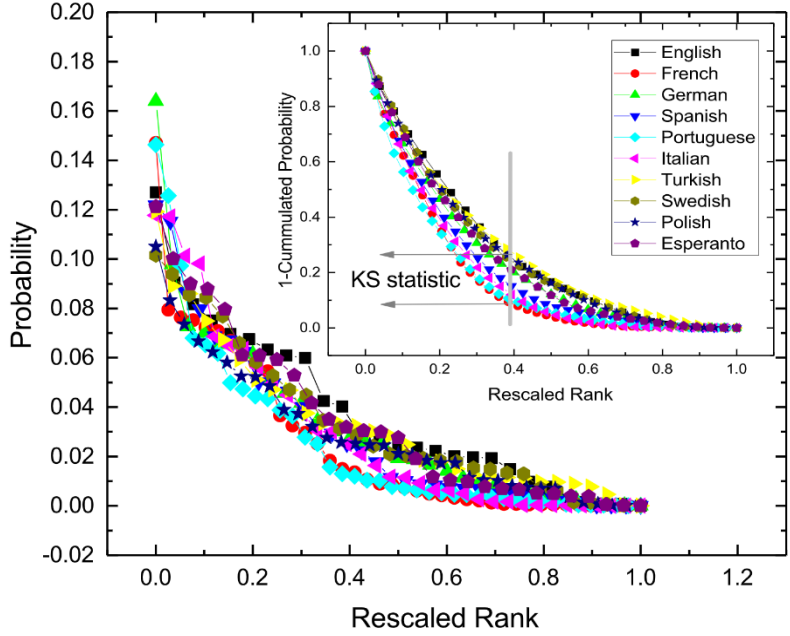

**Figure 2.** Inverse cumulative distribution for 10 alphabetical writing languages.

In the upper-right subplot of Figure 2, we show the inverse cumulative distribution functions, which become smoother in shape. Furthermore, to more precisely quantify the similarities of the letter distributions, we calculated the Kolmogorov-Smirnov (KS) statistics (Drew, Glen, & Leemis, 2000), the maximum absolute value of two cumulative distribution functions with equation as

$$D = \max_r |F(r) - P(r)|. \qquad (1)$$

Here value of $D$ or value of KS statistics is between 0 and 1, and the smaller the $D$ value, the smaller the difference between the two-frequency series, and resulted in the increase of their consistency. Intuitively, we can draw a vertical line to intersect the cumulative distribution function curve of the distribution that we want to compare in Figure 2. The $D$ values of each pair in the 10 languages are displayed in Table 3.

Among the table, the minimum value was 0.034, indicating that the letter usages in Swedish and Turkish was the most similar. In contrast, the letter usages in English and French were the most dissimilar due to the $D$ is 0.226. However, the mean of the $D$ equaling 0.137 implies that the distance between any two languages is not significantly different. Among the examined languages, Portuguese is certainly different from others. In this case, the average $D$ is 0.175 comparing to other languages; if Portuguese is removed from consideration, the average $D$ among all language decreases to 0.133. Among the 10 languages, Polish and Esperanto have minimum average $D$ of 0.120 and 0.125 comparing to all other languages.

**Table 3.** *D* value of the letter frequency distributions for 10 alphabetical languages.

|            | French | German | Spanish | Portuguese | Italian | Turkish | Swedish | Polish | Esperanto |
|------------|--------|--------|---------|------------|---------|---------|---------|--------|-----------|
| English    | 0.226  | 0.135  | 0.165   | 0.218      | 0.194   | 0.081   | 0.068   | 0.075  | 0.104     |
| French     | 0      | 0.167  | 0.112   | 0.066      | 0.109   | 0.222   | 0.201   | 0.210  | 0.173     |
| German     | -      | 0      | 0.080   | 0.160      | 0.111   | 0.143   | 0.160   | 0.095  | 0.140     |
| Spanish    | -      | -      | 0       | 0.150      | 0.036   | 0.154   | 0.136   | 0.111  | 0.116     |
| Portuguese | -      | -      | -       | 0          | 0.154   | 0.212   | 0.222   | 0.205  | 0.191     |
| Italian    | -      | -      | -       | -          | 0       | 0.188   | 0.163   | 0.142  | 0.148     |
| Turkish    | -      | -      | -       | -          | -       | 0       | 0.034   | 0.069  | 0.087     |
| Swedish    | -      | -      | -       | -          | -       | -       | 0       | 0.087  | 0.074     |
| Polish     | -      | -      | -       | -          | -       | -       | -       | 0      | 0.088     |

### 3.2. Fitting of the letter distributions

Similarly, referring the Zipf's law, the change in the relative frequency of letters with a rank can also be analysed. In this paper, we refer to the 7 most popular equations (Deng, 2016; Grzybek, 2007; Grzybek & Rusko, 2009; Li & Miramontes, 2011) to fit the letter distribution curve.

**Table 4.** Fitting results of different equations for letter distribution in English.

| No. | Equation | $\hat{A}$ | $\hat{a}$ | $\hat{b}$ | $R^2$ | RSS | $AIC_c$ |
|-----|----------|-----------|-----------|-----------|-------|------|---------|
| 1 | $p(r) = A r^{-a}$ | 0.142 (0.0119) | 0.599 (0.0537) | - | 0.830 | 0.0046 | -220.1 |
| 2 | $p(r) = A e^{-ar}$ | 0.128 (0.0050) | 0.112 (0.0061) | - | 0.962 | 0.0010 | -259.2 |
| 3 | $p(r) = A - a\log(r)$ | 0.129 (0.0036) | 0.039 (0.0015) | - | 0.967 | 0.0009 | -262.8 |
| 4 | $p(r) = A - a\log(r) - b[\log(r)]^2$ | 0.120 (0.0046) | 0.024 (0.0050) | 0.004 (0.0013) | 0.976 | 0.0006 | -268.6 |
| 5 | $p(r) = A r^{-a} e^{\frac{-b}{r}}$ | 0.411 (0.1224) | 1.002 (0.1271) | 1.238 (0.3429) | 0.897 | 0.0028 | -230.5 |
| 6 | $p(r) = A r^{-a}(n + 1 - r)^b$ | 0.002 (0.0010) | 0.202 (0.0416) | 1.280 (0.1533) | 0.978 | 0.0006 | -271.1 |
| 7 | $\dfrac{\binom{a+r-2}{r-1}\binom{b-a+A-r}{A-r+1}}{\binom{b+A-1}{A}}$ | 39 | 1 | 5.766 | 0.965 | 0.0009 | -258.9 |

() represents the standard deviation for parameter estimation.
We find the optimal solution of equation (7) the Negative Hypergeometric Distribution (NHG) by searching space, so there is no standard deviation.
The $AIC_c = n \ln\left(\frac{RSS}{n}\right) + 2k + \frac{2k(k+1)}{n-k-1}$ from reference (Deng, 2016), there *n* is the amount of points and k is number of parameters.

As shown in Table 4, where $p(r)$ refers to the frequency or probability of the *r*-th frequently used letter, *n* is the total number of letters. As observed from the results of the fitting, e.g., the coefficient of determination ($R^2$) and the residual sum of squares (*RSS*), a larger $R^2$ and smaller *RSS* indicate a better fit. Here, $AIC_c$ can also indicate the quality of the estimation, but it takes into account the influence of degrees of freedom. Lower $AIC_c$ means better fitting effect. From the data in the table, we can see that these equations are very good for fitting English letters, which confirms Li's result (Li & Miramontes, 2011). More detail, The ($R^2$) all go beyond 0.96 except the fittings for $p(r) = A r^{-a}$ and $p(r) = A r^{-a} e^{-b/r}$. The Cocho /Beta equation $p(r) = A r^{-a}(n + 1 - r)^b$ and quadratic logarithmic function $p(r) = A - a\log(r) - b[\log(r)]^2$ perform best in this case. These equations can also obtain good results when applied to the letter distributions in other languages with the smaller sum of squares of residuals and larger determinant coefficients.

## 4. The 'letter' frequency in Chinese

The consistency of the various alphabetical languages leads to the following question: does the Chinese language have the same features as those of other languages? This question is not easy to answer because there are no explicit letters in Chinese. The first question that we must answer is: what is the counterpart to letters in Chinese? We discuss and analyze some possible options in this section. They are Chinese characters, Chinese strokes and Chinese constructive parts.

### 4.1. *The Chinese characters*

Sentences in Chinese are composed of a series of arranged characters, which represent the most natural unit (Wong, Li, Xu, & Zhang, 2009). For example, "我们爱和平" includes 5 characters "我", "们", "爱", "和", and "平". In this sense, someone may think that Chinese characters, the natural units in sentences, should be equated to alphabetical letters. To determine whether this is true, we compared the frequencies of letters and Chinese characters. However, there are more than 7,000 types of Chinese characters, whereas there are only approximately 30 letters for English and other languages. Thus, their distributions are located in completely different ranges. To compare the two distributions more reasonably, we re-scaled the ranking to fall in the range of 0 to 1. Then, we compared the resulting distribution with the distributions of letters in the 10 alphabetical writing languages, as shown in Figure 3. The inverse cumulative distribution of re-scaled Chinese characters is steeper than the ten languages we explained earlier, which indicates that there is a greater imbalance in the use of Chinese characters. The KS distance between Chinese character distributions during different periods are relatively small even though the most frequent characters are quite different (Q. Chen et al., 2012), the average *D* value is only 0.105, especially between Corpus 3 and Corpus 4 that is only 0.019.

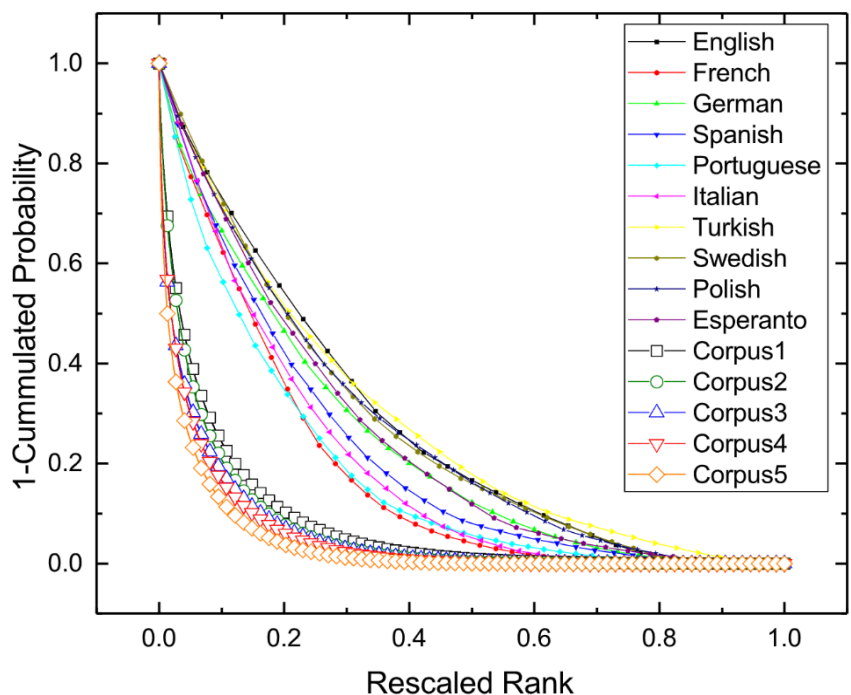

**Figure 3.** Letter distribution with Chinese characters as letters.

However, the KS statistics between Chinese and the 10 alphabetical writing languages ranged from 0.327 to 0.618 with an average of 0.499. The significant gap between Chinese character distributions and the letter distributions implies that Chinese characters could not be considered counterpart to the letter in other languages.

### 4.2. *The Chinese strokes*

Chinese strokes are another candidate counterparts for letters because they compose Chinese characters. Strokes are an attempt to identify and classify all single-stroke components that can be used to write Han radicals. There are some distinct types of strokes recognized in Chinese characters, some of which are compound strokes made from basic strokes. The compound strokes comprise more than one movement of the writing instrument. Different scholars have different interpretations. For example, R. Chen and Chen (1998) analyzed 1000 high-frequency Chinese characters and concluded that there are 6 basic strokes and 22 derivative strokes. Huang and Liao (1997) edited "the Modern Chinese (the second edition)", in which they stated that there exist 5 basic strokes and 36 extended strokes.

We collected statistical data of strokes of Chinese characters from https://github.com/DongSky/zhHanSequence. In these data set, each character is determined to be derived from 24 basic strokes, which is almost the same as the number of alphabetical language letters. The strokes and some example characters are shown in Figure 4.

**Figure 4.** Some Chinese Strokes.

After computing, we compared their frequency distributions with letters in a rescaling scaled plot, as shown in Figure 5.

The frequency of Chinese strokes is very close, and the average KS statistics of Chinese stroke distribution for 5 corpora is 0.010. The maximum is 0.0175, which is the distance between corpus 2 and corpus 5, and the minimum is 0.003, which is the distance between corpus 3 and corpus 4. The mean of *D* from the Chinese strokes to the 10 alphabetical languages is 0.340. The closest language to the average distance of the distribution of Chinese strokes is Portuguese, and the average KS statistics is 0.247. The farthest is Swedish, and the average KS statistics to Chinese strokes is 0.384. Although this KS statistic is better than the comparison between Chinese characters and English, they all exceed the *D* values among the 10 alphabetical languages. Therefore, we concluded that strokes are inconsistent with letters for this analysis.

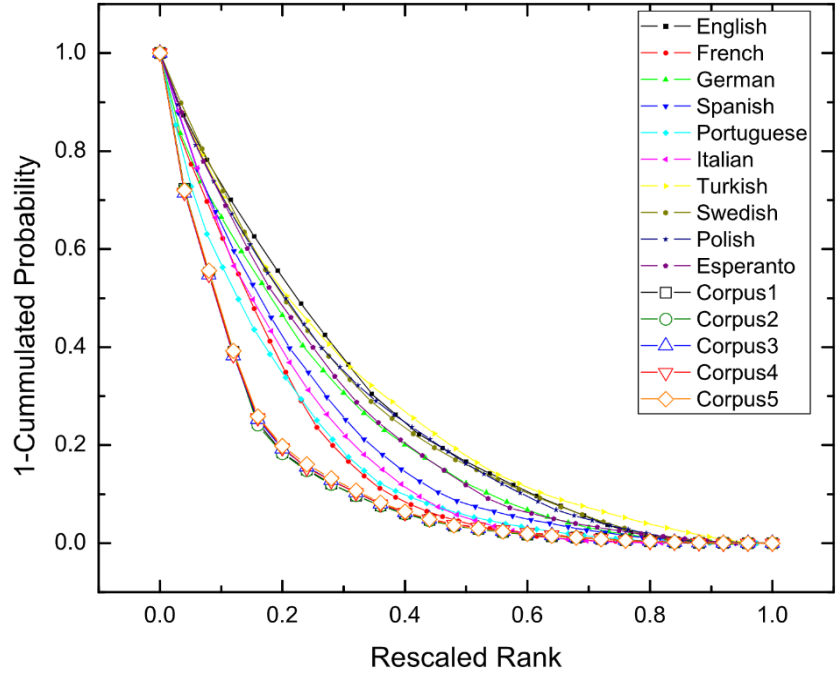

**Figure 5.** Frequency of letter with the strokes of Chinese characters.

### 4.3. *Chinese character constructive parts*

After the above analysis, we found that there are large differences between the frequencies of letters and Chinese characters or strokes. We are convinced that letters and characters or strokes do not belong to the same hierarchy. In general, words are regarded as the carrier of meaning. However, most characters in Chinese can express a clear meaning (Wang, Li, & Di, 2005). If words are split into basic strokes, the strokes cannot express the meaning of the words. For English and some other languages, the roots of the word always have a specific meaning. Even a single letter can express partial meaning of the word sometimes. Hence, we assume that the complexity level of the word is too high. On the contrary, the complexity level of the stroke is too low, and the "letters" of Chinese characters should be in the middle tier of the characters and the strokes.

Jinchang Fei, a Chinese linguist, considers that the Chinese character constructive part can form an independent character, which is greater than or equal to the stroke and less than or equal to the whole character. However, there are various decomposing methods from different schools of thought (Fei, 1996). For example, there are 560 kinds of parts in "Chinese Character Component Standard of GB 13000.1 Character Set for Information Processing" which is formulated under the joint chairmanship of the State Language Commission and the Press and Publication Administration of China. However, some scholars decomposed the most-used 3500 characters into 239 sub-characters (Yan, Fan, Di, Havlin, & Wu, 2013).

Mr. Mu Li has been engaged in the study of the structure of the Chinese language for a long time, including a comparative study of simplified and traditional Chinese characters. He summarised the various decomposition methods for Chinese character constructive parts and proposed a new reasonable scheme for a comprehensive consideration of formation and determining meaning. He divided 7118 commonly used simplified Chinese characters into approximately 340 parts and posted the result on the website http://chinese.exponode.com/0_1.htm. For example, "皑" is composed by 5 parts "白山一口兰" and "暗" by 3 parts "日立日".

**Table 5.** Usage frequencies and probabilities (%) of Chinese constructive parts for different corpora.

| | Corpus 1 | | | Corpus 2 | | | Corpus 3 | | | Corpus 4 | | | Corpus 5 | | |
|---|---|---|---|---|---|---|---|---|---|---|---|---|---|---|---|
| Rank | Cons. | Freq. | Prob.% | Cons. | Freq. | Prob.% | Cons. | Freq. | Prob.% | Cons. | Freq. | Prob.% | Cons. | Freq. | Prob.% |
| 1 | 口 | 380898 | 5.828 | 口 | 193201 | 5.339 | 口 | 287767 | 5.539 | 口 | 348427 | 5.852 | 口 | 1663932 | 5.574 |
| 2 | 一 | 248042 | 3.795 | 一 | 134451 | 3.716 | 一 | 260223 | 5.009 | 一 | 290605 | 4.881 | 一 | 1478423 | 4.953 |
| 3 | 日 | 200607 | 3.069 | 日 | 119481 | 3.302 | 日 | 144684 | 2.785 | 土 | 152824 | 2.567 | 土 | 829558 | 2.779 |
| 4 | 木 | 145444 | 2.225 | 木 | 89680 | 2.478 | 亻 | 133097 | 2.562 | 日 | 148211 | 2.489 | 日 | 801085 | 2.684 |
| 5 | 氵 | 137057 | 2.097 | 氵 | 79091 | 2.186 | 土 | 109931 | 2.116 | 亻 | 127376 | 2.140 | 丶 | 704781 | 2.361 |
| 6 | 十 | 136637 | 2.091 | 十 | 74207 | 2.051 | 厶 | 108940 | 2.097 | 人 | 126088 | 2.118 | 亻 | 640194 | 2.145 |
| 7 | 土 | 130175 | 1.992 | 亻 | 71587 | 1.978 | 木 | 101392 | 1.952 | 丶 | 110653 | 1.859 | 人 | 571391 | 1.914 |
| 8 | 月 | 119566 | 1.829 | 月 | 69228 | 1.913 | 十 | 96748 | 1.862 | 辶 | 95650 | 1.607 | 辶 | 512771 | 1.718 |
| 9 | 亻 | 111969 | 1.713 | 土 | 67682 | 1.870 | 人 | 92362 | 1.778 | 木 | 93865 | 1.577 | 丿 | 510993 | 1.712 |
| 10 | 人 | 108520 | 1.660 | 人 | 66949 | 1.850 | 丿 | 88230 | 1.698 | 丿 | 90646 | 1.523 | 白 | 505958 | 1.695 |
| 11 | 八 | 98719 | 1.510 | 卄 | 55359 | 1.530 | 小 | 87969 | 1.693 | 厶 | 86861 | 1.459 | 月 | 505386 | 1.693 |
| 12 | 卄 | 92978 | 1.423 | 小 | 53358 | 1.475 | 二 | 81061 | 1.560 | 十 | 85152 | 1.430 | 厶 | 488266 | 1.636 |
| 13 | 宀 | 85014 | 1.301 | 大 | 51359 | 1.419 | 丶 | 78635 | 1.514 | 小 | 84339 | 1.417 | 勹 | 481947 | 1.615 |
| 14 | 小 | 84759 | 1.297 | 八 | 50482 | 1.395 | 扌 | 77739 | 1.496 | 大 | 83378 | 1.400 | 小 | 444349 | 1.489 |
| 15 | 大 | 83017 | 1.270 | 宀 | 48712 | 1.346 | 月 | 70457 | 1.356 | 八 | 83097 | 1.396 | 木 | 431342 | 1.445 |
| 16 | 厶 | 81900 | 1.253 | 又 | 45647 | 1.261 | 宀 | 62830 | 1.209 | 月 | 81197 | 1.364 | 宀 | 431020 | 1.444 |
| 17 | 又 | 81309 | 1.244 | 丿 | 45142 | 1.248 | 乂 | 62191 | 1.197 | 扌 | 80481 | 1.352 | 十 | 412003 | 1.380 |
| 18 | 丿 | 79791 | 1.221 | 丶 | 43748 | 1.209 | 大 | 61243 | 1.179 | 丷 | 77107 | 1.295 | 扌 | 411773 | 1.379 |
| 19 | 宀 | 78413 | 1.200 | 厶 | 42091 | 1.163 | 女 | 60972 | 1.174 | 女 | 71769 | 1.205 | 丨 | 402468 | 1.348 |
| 20 | 丶 | 77516 | 1.186 | 宀 | 41203 | 1.139 | 辶 | 57353 | 1.104 | 寸 | 71266 | 1.197 | 大 | 366259 | 1.227 |
| 21 | 匕 | 70356 | 1.076 | 匕 | 39065 | 1.080 | 八 | 57021 | 1.098 | 丨 | 70217 | 1.179 | 又 | 354400 | 1.187 |
| 22 | 辶 | 66465 | 1.017 | 寸 | 33577 | 0.928 | 又 | 54437 | 1.048 | 又 | 70112 | 1.178 | 寸 | 328515 | 1.101 |
| 23 | 寸 | 62249 | 0.952 | 氵 | 33398 | 0.923 | 寸 | 54017 | 1.040 | 乂 | 65256 | 1.096 | 疋 | 322904 | 1.082 |
| 24 | 夂 | 59111 | 0.904 | 二 | 32429 | 0.896 | 丨 | 53978 | 1.039 | 宀 | 63534 | 1.067 | 也 | 317489 | 1.064 |
| 25 | 冂 | 58893 | 0.901 | 女 | 31724 | 0.877 | 儿 | 53868 | 1.037 | 匕 | 61434 | 1.032 | 丷 | 317162 | 1.063 |
| 26 | 二 | 58092 | 0.889 | 冂 | 31528 | 0.871 | 丷 | 52281 | 1.006 | 宀 | 61399 | 1.031 | 又 | 315540 | 1.057 |
| 27 | 心 | 57082 | 0.873 | 心 | 30679 | 0.848 | 戈 | 51111 | 0.984 | 兰 | 60622 | 1.018 | 左 | 311939 | 1.045 |
| 28 | 冖 | 56443 | 0.864 | 夕 | 30480 | 0.842 | 宀 | 51095 | 0.984 | 儿 | 59823 | 1.005 | 匕 | 296778 | 0.994 |
| 29 | 夕 | 55805 | 0.854 | 丷 | 30343 | 0.839 | 氵 | 49928 | 0.961 | 曰 | 53641 | 0.901 | 辶 | 290051 | 0.972 |
| 30 | 丨 | 52793 | 0.808 | 辶 | 29613 | 0.818 | 夂 | 49494 | 0.953 | 辶 | 50487 | 0.848 | 卜 | 282312 | 0.946 |
| 31 | 丁 | 52078 | 0.797 | 丁 | 29051 | 0.803 | 丁 | 48853 | 0.940 | 二 | 49492 | 0.831 | 儿 | 274685 | 0.920 |
| 32 | 丷 | 51589 | 0.789 | 冖 | 28092 | 0.776 | 曰 | 47928 | 0.923 | 左 | 48675 | 0.818 | 二 | 269934 | 0.904 |
| 33 | 山 | 48189 | 0.737 | 彐 | 27853 | 0.770 | 匕 | 45090 | 0.868 | 氵 | 47969 | 0.806 | 目 | 267445 | 0.896 |
| 34 | 彐 | 47592 | 0.728 | 丨 | 27346 | 0.756 | 白 | 41858 | 0.806 | 戈 | 47820 | 0.803 | 八 | 264001 | 0.884 |
| 35 | 氵 | 46434 | 0.710 | 纟 | 26810 | 0.741 | 心 | 41839 | 0.805 | 疋 | 47469 | 0.797 | 了 | 256611 | 0.860 |
| 36 | 女 | 45831 | 0.701 | 夂 | 26502 | 0.732 | 卄 | 40929 | 0.788 | 王 | 46901 | 0.788 | 戈 | 245088 | 0.821 |
| 37 | 左 | 43093 | 0.659 | 禾 | 25059 | 0.693 | 子 | 40428 | 0.778 | 不 | 45916 | 0.771 | 不 | 240568 | 0.806 |
| 38 | 儿 | 42918 | 0.657 | 王 | 25031 | 0.692 | 卜 | 39826 | 0.767 | 卜 | 45364 | 0.762 | 女 | 237025 | 0.794 |
| 39 | 兰 | 42311 | 0.647 | 儿 | 24337 | 0.673 | 也 | 39380 | 0.758 | 心 | 44931 | 0.755 | 心 | 234548 | 0.786 |
| 40 | 王 | 42181 | 0.645 | 年 | 24204 | 0.669 | 疋 | 38073 | 0.733 | 自 | 43806 | 0.736 | 宀 | 215987 | 0.724 |
| 41 | 年 | 41812 | 0.640 | 讠 | 23347 | 0.645 | 左 | 37898 | 0.729 | 丁 | 42871 | 0.720 | 氵 | 215538 | 0.722 |
| 42 | 厂 | 40905 | 0.626 | 目 | 23335 | 0.645 | 氵 | 37550 | 0.723 | 白 | 42623 | 0.716 | 兰 | 202526 | 0.678 |
| 43 | 乂 | 40609 | 0.621 | 工 | 23238 | 0.642 | 不 | 36186 | 0.697 | 了 | 42463 | 0.713 | 冂 | 201857 | 0.676 |
| 44 | 扌 | 39297 | 0.601 | 兰 | 21915 | 0.606 | 目 | 35951 | 0.692 | 子 | 41645 | 0.700 | 夂 | 201443 | 0.675 |
| 45 | 戈 | 38969 | 0.596 | 扌 | 21882 | 0.605 | 讠 | 34796 | 0.670 | 冂 | 40774 | 0.685 | 丁 | 188293 | 0.631 |
| 46 | 曰 | 38426 | 0.588 | 山 | 21740 | 0.601 | 贝 | 34075 | 0.656 | 卄 | 39528 | 0.664 | 氵 | 186060 | 0.623 |
| 47 | 尸 | 37857 | 0.579 | 曰 | 21534 | 0.595 | 勹 | 34074 | 0.656 | 冖 | 38956 | 0.654 | 阝 | 181180 | 0.607 |
| 48 | 纟 | 37764 | 0.578 | 火 | 20973 | 0.580 | 冖 | 33976 | 0.654 | 彳 | 38824 | 0.652 | 曰 | 179999 | 0.603 |
| 49 | 禾 | 37688 | 0.577 | 入 | 20118 | 0.556 | 止 | 33937 | 0.653 | 目 | 36667 | 0.616 | 自 | 178936 | 0.599 |
| 50 | 一 | 37571 | 0.575 | 乂 | 20097 | 0.555 | 人 | 33099 | 0.637 | 人 | 36665 | 0.616 | 刂 | 174126 | 0.583 |

*Some constructive parts do not be displayed for font reasons.

Using this method, we split the corpora of different Chinese characters into Chinese character constructive parts. The resulting data are displayed in Table 5. Although different corpora differ in the frequency of characters used, as pointed out in the reference (Q. Chen et al., 2012), the most commonly used characters are "不", "人", "云", "了" and "的" for different corpus respectively. But the most commonly used constructive part is "口" and second one is "一", and the frequencies are fairly consistent, with a probability of 5.339% − 5.852%.

The distribution of Chinese constructive parts is shown in Figure 6. We compared the resulting distribution with the letter distribution and found that the distributions were relatively close.

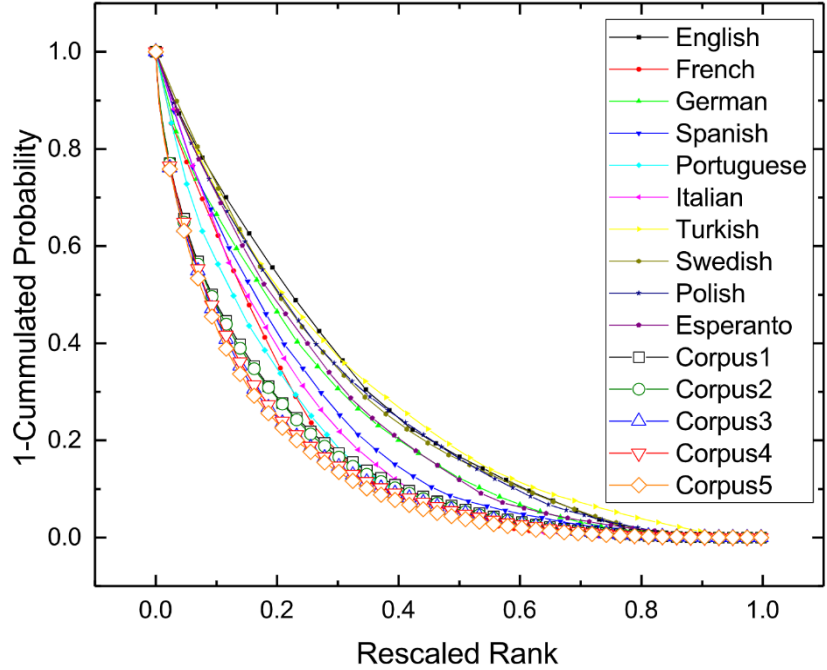

**Figure 6.** Frequency of Chinese character constructive parts.

Furthermore, to explore the similarity of Chinese constructive parts of corpora during different periods, we used the KS statistics to investigate the greatest distance between two cumulative distribution functions. The results are shown in Table 6. The minimum values of the KS statistics of these Chinese corpora compared to the 10 alphabetic languages are less than 0.15, and most of them do not exceed the average KS statistic among the 10 alphabetic languages of 0.137. From these results, it was clear that the distance of 'letter' distribution between Chinese and the 10 alphabetical languages was less. Chinese was found to be the most similar to Portuguese in 'letter' usage.

**Table 6.** KS statistics of the frequency of Chinese character constructive parts for corpora in different periods.

|  | Corpus2 | Corpus3 | Corpus4 | Corpus5 | 10 Alphabetical Language | | |
|---|---|---|---|---|---|---|---|
|  |  |  |  |  | minimum | maximum | average |
| Corpus1 | 0.009 | 0.048 | 0.039 | 0.061 | 0.100 | 0.262 | 0.199 |
| Corpus2 | 0 | 0.043 | 0.036 | 0.056 | 0.103 | 0.268 | 0.204 |
| Corpus3 | - | 0 | 0.013 | 0.019 | 0.124 | 0.309 | 0.235 |
| Corpus4 | - | - | 0 | 0.028 | 0.116 | 0.300 | 0.226 |
| Corpus5 | - | - | - | 0 | 0.143 | 0.323 | 0.251 |

We compare the KS distance between Chinese corpus and other languages. In Figure 7, the hollow blue circles indicate the KS statistics or *D* value among the alphabetical language family, and the solid red circles express the distance between Chinese corpus and other 10 alphabetical languages. Comparing to the hollow and solid circles, the solid circles represent a larger KS statistics as $D \in [0.013, 0.061]$, which means the Chinese corpus has relatively larger distance from alphabetical languages, but multiple red solid circles have fallen into the distance range of hollow blue circles, which indicates that the frequency of Chinese character constructive parts have the relatively similar distribution with the alphabetical language family.

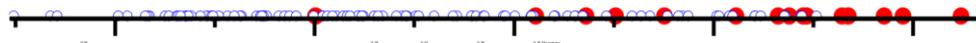

**Figure 7.** The KS statistics among Chinese corpus 1 and other language. Red circle represents the KS statistics of Chinese corpus 1 to 10 alphabetical languages.

To compare these 'letter' data more convenient, we make 31 groups by combining sequential 11 Chinese constructive parts. We get 31 Chinese representative 'letters'. The distributions of Chinese 'letter' and the letters of 10 languages are shown in Figure 8.

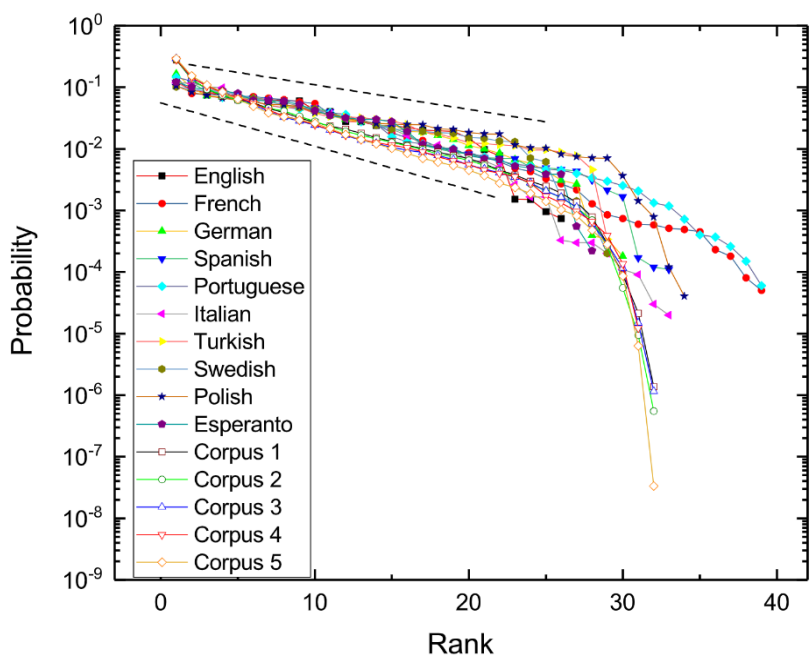

**Figure 8.** Frequency of usage of Chinese character components and letters in the 10 alphabetical language. The vertical axis is in logarithmic scale.

The horizontal axis of the graph is letter ranking on a linear scale, and the longitudinal axis is the probability presented on logarithmic coordinates. The dashed lines indicate that these distributions have strong exponential distribution characteristics in the ranking intervals ranging from 3 to 25. The slopes of the "Chinese letter" distribution are larger compared to the other evaluated distributions which signifies that the intensity of use in Chinese is more imbalanced than the others.

To explore the specific form of the distribution functions, we used 7 different equations to fit the letter distribution curve. The results are shown in Table 7 and Table 8. Obviously, there are large $R^2$ and small $RSS$, which imply a good fit by the functions and indicate that they are suitable for describing the letter distribution. With the exception of the fit of the logarithmic function $p(r) = A − a\log(r)$, the remaining functions have ($R^2$) values above 0.91, and these functions produce results that are even better than those obtained for English. In these cases, the Cocho/Beta equation, quadratic logarithmic equation, and Inverse-Gamma function $p(r) = Ar^{-a}e^{\frac{-b}{r}}$ fit best.

Table 7. Fitting results of different equations for Chinese character constructive parts of Chinese corpus

| No. | Equation | $\hat{A}$ | $\hat{a}$ | $\hat{b}$ | $R^2$ | RSS | $AIC_c$ |
|---|---|---|---|---|---|---|---|
| 1 | $p(r) = Ar^{-a}$ | 0.282 (0.0061) | 1.044 (0.0254) | - | 0.987 | 0.0013 | -320.3 |
| 2 | $p(r) = Ae^{-ar}$ | 0.345 (0.0250) | 0.349 (0.0298) | - | 0.933 | 0.0062 | -269.2 |
| 3 | $p(r) = A − a\log(r)$ | 0.179 (0.0133) | 0.0579 (0.0049) | - | 0.821 | 0.0166 | -237.6 |
| 4 | $p(r) = A − a\log(r) − b[\log(r)]^2$ | 0.427 (0.0669) | 1.230 (0.0770) | 0.438 (0.1621) | 0.989 | 0.0010 | -326.2 |
| 5 | $p(r) = Ar^{-a}e^{\frac{-b}{r}}$ | 0.0075 (0.0042) | 0.874 (0.0236) | 1.039 (0.1616) | 0.997 | 0.0003 | -364.9 |
| 6 | $p(r) = Ar^{-a}(n + 1 − r)^b$ | 0.256 (0.0068) | 0.158 (0.0068) | -0.0248 (0.0016) | 0.980 | 0.0019 | -305.3 |
| 7 | $\frac{\binom{a+r-2}{r-1}\binom{b-a+A-r}{A-r+1}}{\binom{b+A-1}{A}}$ | 984 | 1 | 306.7 | 0.915 | 0.0079 | -258.9 |

[a]()represents the standard deviation for parameter estimation.

Table 8. Fitting results of different equations for Chinese character constructive parts of Chinese corpus

| No. | Equation | $\hat{A}$ | $\hat{a}$ | $\hat{b}$ | $R^2$ | RSS | $AIC_c$ |
|---|---|---|---|---|---|---|---|
| 1 | $p(r) = Ar^{-a}$ | 0.302 (0.0077) | 1.066 (0.0315) | - | 0.982 | 0.0019 | -305.6 |
| 2 | $p(r) = Ae^{-ar}$ | 0.384 (0.0210) | 0.368 (0.0234) | - | 0.966 | 0.0037 | -285.7 |
| 3 | $p(r) = A − a\log(r)$ | 0.191 (0.0148) | 0.0625 (0.0056) | - | 0.818 | 0.0198 | -232.0 |
| 4 | $p(r) = A − a\log(r) − b[\log(r)]^2$ | 0.278 (0.0045) | 0.176 (0.0047) | -0.028 (0.0011) | 0.993 | 0.0008 | -331.8 |
| 5 | $p(r) = Ar^{-a}e^{\frac{-b}{r}}$ | 0.724 (0.0988) | 1.471 (0.0720) | 0.915 (0.1436) | 0.993 | 0.0007 | -335.6 |
| 6 | $p(r) = Ar^a(n + 1 − r)^b$ | 0.00025 (0.0002) | 0.788 (0.0130) | 2.032 (0.1101) | 0.999 | 0.0001 | -408.7 |
| 7 | $\frac{\binom{a+r-2}{r-1}\binom{b-a+A-r}{A-r+1}}{\binom{b+A-1}{A}}$ | 899 | 1 | 324.4 | 0.954 | 0.0049 | -273.7 |

5. [a]()represents the standard deviation for parameter estimation.

Through the above analysis, we can see different alphabetical writing languages with similar frequency distribution characteristics. Similarly, the Chinese constructive parts in different historical periods have almost the same distribution. Both distributions were well fit with the Cocho/Beta Equation $p(r) = Ar^{-a}(n + 1 − r)^b$ and quadratic logarithmic equation $p(r) = A − a\log(r) − b[\log(r)^2]$.

## 5. Conclusions and Discussion

Linguistic studies have gradually revealed that different languages have more and more common characteristics. While they may not formally look the same, statistical analysis shows that many languages are strongly consistent with each other in terms of word frequency, semantic structure, grammatical network, and even in regard to bias. Exploring language consistency is fascinating but challenging work.

In this paper, we focus on the frequency of letter use. We discuss the distribution for letter frequency of 10 alphabetical writing languages and find the letter distributions to be very consistent. Additionally, we conducted a statistical analysis involves the corpora of Chinese literature throughout different historical periods from the Tang Dynasty to the present. We found the Chinese constructive parts of having similar statistics to characters with letters in other languages. The data could be well-fitted by the same equations, which is significant evidence.

As yet, there is no standard way to decompose the Chinese characters into more basic components. Different scholars have different ways of decomposing them. For example, Yan et al. (2013) decomposed the most-used 3500 characters into 239 parts. We also analysed these data and found the results to be similar. The distributions can also be well-fit using the 7 equations. However, there is more inconsistency in the use of components. The mean KS value among Chinese corpora is $0.026$, and the KS statistic among Chinese corpora and 10 other languages is slightly larger, with a value of $0.272$.

The distributions are surprisingly consistent among various languages, which reflects the natural consistency of human language. Different countries and nationalities in the world have different languages, but they have common uses and purpose so that more consistencies shall be found and examined. In addition to the distributional analyses, we also should try to discover the properties of the organization of these units in their syntagmatic dimension (Köhler, 2008). This integration process can enable human beings to have a deeper understanding of language and make better use of it, for example through machine translation, machine writing, and even contributing to the exploration and utilization of the human brain.

**Acknowledgement**

We thank Professors Zengru Di and Yougui Wang for their discussions and suggestions. This work was supported by the National Social Science Foundation of China (Grant Nos. 71701018 and 61673070) and China Scholarship Council.

# References

Ausloos, M. (2010). Punctuation effects in english and esperanto texts. *Physica A: Statistical Mechanics and its Applications*, *389*(14), 2835–2840.

Beliankou, A., Köhler, R., & Naumann, S. (2012). Quantitative properties of argumentation motifs. In *Methods and applications of quantitative linguistics, selected papers of the 8th international conference on quantitative linguistics (qualico)* (pp. 35–43).

Brown, P. F., Desouza, P. V., Mercer, R. L., Pietra, V. J. D., & Lai, J. C. (1992). Class-based n-gram models of natural language. *Computational Linguistics*, *18*(4), 467–479.

Chen, Q., Guo, J., & Liu, Y. (2012). A statistical study on chinese word and character usage in literatures from the tang dynasty to the present. *Journal of Quantitative Linguistics*, *19*(3), 232–248.

Chen, R., & Chen, A. (1998). Analysis and teaching conception of 1000-frequency chinese characters (in chinese). *Language and Character Application*(1), 49–53.

Choi, S.-W. (2000). Some statistical properties and zipf's law in korean text corpus. *Journal of Quantitative linguistics*, *7*(1), 19–30.

Dalkılıç, G., & Çebi, Y. (2004). Zipf's law and mandelbrot's constants for turkish language using turkish corpus (turco). In *International conference on advances in information systems* (pp. 273–282).

Deng, Y. (2016). Some statistical properties of phonemes in standard chinese. *Journal of Quantitative Linguistics*, *23*(1), 30–48.

Dodds, P. S., Clark, E. M., Desu, S., Frank, M. R., Reagan, A. J., Williams, J. R., … others (2015). Human language reveals a universal positivity bias. *Proceedings of the National Academy of Sciences*, *112*(8), 2389–2394.

Drew, J. H., Glen, A. G., & Leemis, L. M. (2000). Computing the cumulative distribution function of the kolmogorov–smirnov statistic. *Computational Statistics & Data Analysis*, *34*(1), 1–15.

Fei, J. (1996). Exploration of modern chinese character components (in chinese). *Language and Character Application*(2), 20–26.

Grzybek, P. (2007). On the systematic and system-based study of grapheme frequencies: a re-analysis of german letter frequencies. *Glottometrics*, *15*, 82–91.

Grzybek, P., & Rusko, M. (2009). Letter, grapheme and (allo-) phone frequencies: the case of slovak. *Glottotheory*, *2*(1), 30–48.

Ha, L. Q., Stewart, D. W., Hanna, P., & Smith, F. J. (2006). Zipf and type-token rules for the english, spanish, irish and latin languages. *Web Journal of Formal, Computational and Cognitive Linguistics*, *1*(8), 1–12.

Hatzigeorgiu, N., Mikros, G., & Carayannis, G. (2001). Word length, word frequencies and zipf's law in the greek language. *Journal of Quantitative Linguistics*, *8*(3), 175–185.

Heaps, H. S. (1978). *Information retrieval, computational and theoretical aspects*. Academic Press.

Huang, B., & Liao, X. (1997). *Modern chinese (second edition) (in chinese)*. Higher Education Press.

i Cancho, R. F., Solé, R. V., & Köhler, R. (2004). Patterns in syntactic dependency networks. *Physical Review E*, *69*(5), 051915.

Jaskiewicz, G. (2011). Analysis of letter frequency distribution in the voynich manuscript. In *International workshop on concurrency, specification and programming (cs&p'11)* (pp. 250–261).

Jayaram, B., & Vidya, M. (2008). Zipf's law for indian languages. *Journal of Quantitative Linguistics*, *15*(4), 293–317.

Jernigan, R. W. (2008). A photographic view of cumulative distribution functions. *J. Stat. Educ*, *16*.

Jing, Y., & Liu, H. (2017). Dependency distance motifs in 21 indoeuropean langauges. *Motifs in Language and Text*, 133–150.

Köhler, R. (2008). Sequences of linguistic quantities report on a new unit of investigation. *Glottotheory*, *1*(1), 115–119.

Li, W., & Miramontes, P. (2011). Fitting ranked english and spanish letter frequency distribution in us and mexican presidential speeches. *Journal of Quantitative Linguistics*, *18*(4), 359–380.

Liu, H. (2008). The complexity of chinese syntactic dependency networks. *Physica A: Statistical Mechanics and its Applications*, *387*(12), 3048–3058.

Liu, H. (2018). Language as a human-driven complex adaptive system. *Physics of Life Reviews*, *26*(1), 149–151.

Manning, C. D., Manning, C. D., & Schütze, H. (1999). *Foundations of statistical natural language processing*. MIT press.

Masrai, A., & Milton, J. (2016). How different is arabic from other languages? the relationship between word frequency and lexical coverage. *Journal of Applied Linguistics and Language Research*, *3*(1), 15–35.

Piantadosi, S. T. (2014). Zipf's word frequency law in natural language: A critical review and future directions. *Psychonomic Bulletin & Review*, *21*(5), 1112–1130.

Serengil, S. I., & Akin, M. (2011). Attacking turkish texts encrypted by homophonic cipher. In *Proceedings of the 10th wseas international conference on electronics, hardware.*

Wang, D., Li, M., & Di, Z. (2005). True reason for zipf's law in language. *Physica A: Statistical Mechanics and its Applications*, *358*(2), 545–550.

Wiegand, M., Nadarajah, S., & Si, Y. (2018). Word frequencies: A comparison of pareto type distributions. *Physics Letters A*, *382*(9), 621–632.

Wong, K.-F., Li, W., Xu, R., & Zhang, Z.-S. (2009). Introduction to chinese natural language processing. *Synthesis Lectures on Human Language Technologies*, *2*(1), 1–148.

Yan, X., Fan, Y., Di, Z., Havlin, S., & Wu, J. (2013). Efficient learning strategy of chinese characters based on network approach. *PloS One*, *8*(8), e69745.

Youn, H., Sutton, L., Smith, E., Moore, C., Wilkins, J. F., Maddieson, I., … Bhattacharya, T. (2016). On the universal structure of human lexical semantics. *Proceedings of the National Academy of Sciences*, *113*(7), 1766–1771.

Zhang, D., Xu, H., Su, Z., & Xu, Y. (2015). Chinese comments sentiment classification based on word2vec and svm perf. *Expert Systems with Applications*, *42*(4), 1857–1863.

Zipf, G. K. (1949). *Human behavior and the principle of least effort: an introduction to human ecology*. Addison-Wesley.